\documentclass[11pt]{article}
\usepackage{graphicx}

\usepackage[margin=1in]{geometry}
\usepackage{amsmath, amssymb, amsthm}
\newtheorem{definition}{Definition}
\usepackage{graphicx}
\usepackage{booktabs}
\usepackage{float}
\usepackage{subcaption}
\usepackage{url}
\usepackage[round,authoryear,sort&compress]{natbib}
\usepackage{hyperref}  

\title{Estimating the Effective Rank of Vision Transformers via Low-Rank Factorization}
\author{Liyu Zerihun\\ \texttt{liyulg0@gmail.com}}
\date{October 2025}

\begin{document}

\maketitle

\begin{abstract}
Deep networks are heavily over-parameterized, yet their learned representations 
often admit low-rank structure. We introduce a framework for estimating a model's 
intrinsic dimensionality by treating learned representations as projections onto 
a low-rank subspace of the model's full capacity. Our approach: train a full-rank 
teacher, factorize its weights at multiple ranks, and train each factorized student 
via distillation to measure performance as a function of rank.

We define \emph{effective rank} as a \emph{region}, not a point: the smallest 
contiguous set of ranks for which the student reaches 85--95\% of teacher accuracy. 
To stabilize estimates, we fit accuracy vs.\ rank with a monotone PCHIP interpolant 
and identify crossings of the normalized curve. We also define the \emph{effective 
knee} as the rank maximizing perpendicular distance between the smoothed accuracy 
curve and its endpoint secant; an intrinsic indicator of where marginal gains concentrate.

On ViT-B/32 fine-tuned on CIFAR-100 (one seed, due to compute constraints), 
factorizing linear blocks and training with distillation yields an effective-rank 
region of $\approx[16,\,34]$ and an effective knee at $r^\star\approx31$. At rank 
32, the student attains 69.46\% top-1 accuracy vs.\ 73.35\% for the teacher 
($\sim$94.7\% of baseline) while achieving substantial parameter compression. We 
provide a framework to estimate effective-rank regions and knees across architectures 
and datasets, offering a practical tool for characterizing the intrinsic dimensionality 
of deep models.
\end{abstract}

\section{Introduction}

Neural networks today achieve remarkable performance while being massively over-parameterized. Yet many studies have shown that one can reduce the rank of model weights and still reach similar accuracy to the base model \citep{hu2021lora, hao2023lrc, chen2025tunecomp}. Others have argued that deep networks carry an implicit simplicity bias \citep{huh2023simplicity}. All of these ideas point to the same observation: neural networks often converge to lower-dimensional solutions, even when their parameter space is enormous.

The double descent phenomenon \citep{nakkiran2021deep} adds another layer of mystery. In the over-parameterized regime, performance keeps improving as we increase model size. On the one hand, networks collapse to simpler representations; on the other hand, smaller models trained from scratch perform worse. This gap hints that dimensionality is not about absolute model size, but about how that capacity is used.

This paper gives a way to estimate the dimensionality of that lower-dimensional space, what we call the \textit{effective rank region}. We study one architecture, Vision Transformer (ViT-B/32) on CIFAR-100, and use matrix factorization combined with blockwise geometric distillation \citep{hinton2015distilling} to measure where model performance saturates.

We employ geometric distillation, aligning both magnitude and direction of internal 
activations between teacher and student using MSE and cosine losses. While we 
factorize all linear layers and match intermediate block representations, our 
ablation studies (§5) suggest that matching final logits alone may be sufficient 
for rank estimation.

To read off a stable estimate, we fit accuracy vs. rank with a monotone PCHIP curve and define the effective rank region as the smallest contiguous span where the normalized student accuracy sits between 85\% and 95\% of the teacher. We also use a point of 'effective knee', the rank with the maximum perpendicular distance to the secant through the endpoints of the smoothed curve, as a compact indicator of where the gains are concentrated. 

Empirically, we find that the effective rank of ViT-B/32 lies roughly between ranks 16 and 34. Within this span, the validation accuracy rises quickly; the knee lands at about $r^\star \approx 31$, and by rank 32 the model reaches 69.46\% top-1 versus 73.35\% for the teacher (about 94.7\% of baseline). This region-based view, where capacity ramps up and then levels off, captures the part of the curve that actually matters; it represents a saturation point after which the model gets marginal benefits as rank increases.

\section{Related Work}

There have been numerous works on the effective rank of neural networks in the literature. One established definition of matrix effective rank is:

\begin{definition}
The effective rank \citep{roy2007effective} of a matrix $A \in \mathbb{R}^{m \times n}$ can be described as
\[
\mathrm{erank}(A) = \exp\{ H(p_1, p_2, \ldots, p_Q) \},
\]
where \( H(p_1, p_2, \ldots, p_Q) \) is the (Shannon) entropy given by
\[
H(p_1, p_2, \ldots, p_Q) = -\sum_{k=1}^{Q} p_k \log p_k,
\]
and $p_k = \sigma_k / \sum_j \sigma_j$ are the normalized singular values of $A$.
\end{definition}

This measure captures a continuous representation of effective rank by analyzing the distribution of singular values and thereby quantifying the disorder of the matrix. Intuitively, one can view this as a measurement of the entropy of the matrix: the higher the entropy, the higher the rank. As $A$ is a mapping between two vector spaces, higher disorder requires more information to map between spaces. This maps neatly to the rank of a matrix, as the rank indicates the dimensionality of the subspace that $A$ maps onto.

However, this definition operates at the matrix level; different matrices in the network might converge to different effective ranks over time. In our method we view effective rank globally, constraining most of the matrices in the network to one rank, effectively constraining the whole network to a lower-rank subspace.

Beyond single-matrix definitions, recent work has examined effective rank at the representation level. \citet{huh2023simplicity} finds that deep neural networks have a low-rank simplicity bias. Their results reveal a counterintuitive trend: the deeper the network, the lower the effective rank, hinting that deep neural networks have a strong preference for low-rank manifolds during training.

There have been many works on matrix factorization in deep neural networks. A widely used method is \textbf{LoRA} \citep{hu2021lora}, which fine-tunes large language models using a low-rank addition to the matrices. Another framework, \textbf{Low-Rank Clone (LRC)} \citep{hao2023lrc}, employs learnable low-rank projections to realize soft pruning and behavioral cloning of large models. A related framework, \textbf{TuneComp} \citep{chen2025tunecomp}, jointly fine-tunes and compresses by gradually distilling into a pruned low-rank structure. All of these share the premise that model function lies on a low-dimensional manifold. 

Our framework uses explicit low-rank factorization as an analytical probe to map performance against rank. We differ from these efficiency-focused works by treating rank as a variable of measurement, not optimization.

\section{Methodology and Experiments}

We take a pretrained ViT (ViT-B/32) \citep{dosovitskiy2021vit} and fine-tune it on CIFAR-100. The standard ViT encoder is a stack of identical encoder blocks (LayerNorm $\to$ multihead self-attention $\to$ residual, then LayerNorm $\to$ MLP $\to$ residual). 

\paragraph{Factorized version.}
We create the factorized model by replacing every linear map in self-attention and in the MLP with a rank–$r$ factorized linear layer. Concretely, a weight $W\in\mathbb{R}^{d_{\text{out}}\times d_{\text{in}}}$ is written as
\[
W \approx A B,\qquad A \in \mathbb{R}^{d_{\text{out}}\times r},\; B \in \mathbb{R}^{r\times d_{\text{in}}},\;\; r \ll \min\{d_{\text{in}},d_{\text{out}}\}.
\]
We keep the overall block structure the same (LayerNorms, residuals, heads, etc.); only the linear maps are factorized.

\paragraph{Teacher--student signals (hooks).}
For alignment, we read out the internal block representations from both models. In practice we register forward hooks on each encoder block to capture:
\begin{itemize}
  \item the output of multihead self-attention;
  \item the output of the MLP.
\end{itemize}
Let $h^{T}_{\ell,a}$ and $h^{T}_{\ell,m}$ be the teacher’s attention and MLP outputs at block $\ell$, and $h^{S}_{\ell,a}$, $h^{S}_{\ell,m}$ the student’s. We also keep the final logits $z^{T}$ and $z^{S}$.

\paragraph{Geometric distillation loss.}
We align magnitude and direction with a simple MSE+cosine objective, averaged across blocks:
\[
\mathcal{L}_{\text{blocks}}
= \frac{1}{L}\sum_{\ell=1}^{L}\Big(
\underbrace{\|h^{S}_{\ell,a}-h^{T}_{\ell,a}\|_2^2 + \|h^{S}_{\ell,m}-h^{T}_{\ell,m}\|_2^2}_{\text{MSE}}
\;+\;
\underbrace{(1-\cos(h^{S}_{\ell,a},h^{T}_{\ell,a})) + (1-\cos(h^{S}_{\ell,m},h^{T}_{\ell,m}))}_{\text{cosine}}
\Big),
\]
where $\cos(u,v)=\langle u, v\rangle/(\|u\|_2\,\|v\|_2)$; tensor outputs are averaged over tokens/features before summing over blocks. We also add a standard logit matching term:
\[
\mathcal{L}_{\text{logits}}
= \|z^{S}-z^{T}\|_2^2 + \big(1-\cos(z^{S},z^{T})\big).
\]
We weight cosine and MSE equally. The total loss is
\[
\mathcal{L} \;=\; \mathcal{L}_{\text{blocks}} + \mathcal{L}_{\text{logits}}.
\]

\paragraph{Ranks and readout.}
We sweep the factorization rank $r$ across a grid. For each $r$ we train the student with the geometric loss above. 
We use the ranks
\[
\{2,\,4,\,8,\,16,\,24,\,32,\,48,\,64,\,80,\,96,\,112,\,128,\,160,\,192,\,224,\,256\}.
\]
To report the \emph{effective rank region} and the \emph{effective knee}:
\begin{itemize}
  \item We fit accuracy vs.\ rank with a monotone PCHIP interpolant.
  \item We normalize by the teacher’s accuracy and take the smallest contiguous band where the student reaches $85\%\text{--}95\%$ of the teacher.
  \item We define the knee as the rank that maximizes the perpendicular distance between the smoothed accuracy curve and the secant through the endpoints.
\end{itemize}

\paragraph{MLX.}
We use MLX, a small hypothesis-driven experiment framework we built to parallelize rank sweeps and log metrics; all experiments in this paper can be reproduced with a single command (see repository README).

\paragraph{Ablations.}
We also run the same protocol with pure distillation on logits only as well as MSE + Cosine Distillation on logits only. Hyperparameters and exact training details are in the repository.

\paragraph{Notes.}
We use one seed (compute constraints). Cosine/MSE terms are balanced equally. Hooks are attached to every encoder block for both attention and MLP outputs, and logits are aligned at the end.

\section{Results}

We sweep the rank and read off the curve using a monotone PCHIP fit. The knee (maximum distance to the secant through the endpoints) lands at $r^\star \approx 30.98$. Using the 85–95\% rule, the effective rank region is about $[15.73,\;33.53]$. In that span, accuracy jumps fast and then flattens. At $r=32$, the student reaches $69.46\%$ top-1 accuracy versus $73.35\%$ for the teacher (about $94$–$95\%$ of baseline). The loss drops smoothly with rank.

\begin{figure}[H]
  \centering
  \begin{subfigure}{0.48\textwidth}
    \centering
    \includegraphics[width=\linewidth]{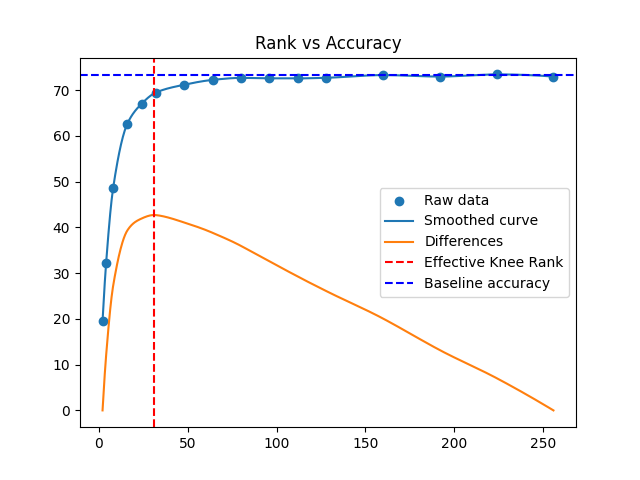}
    \caption{Rank vs.\ Accuracy.}
    \label{fig:rank-acc}
  \end{subfigure}
  \hfill
  \begin{subfigure}{0.48\textwidth}
    \centering
    \includegraphics[width=\linewidth]{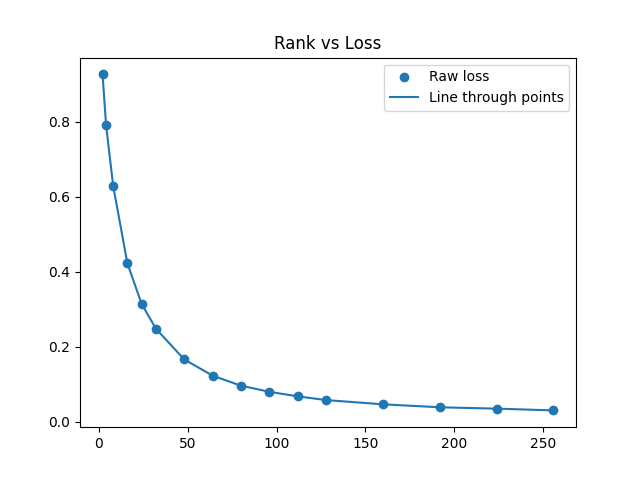}
    \caption{Rank vs.\ Loss.}
    \label{fig:rank-loss}
  \end{subfigure}
  \caption{Model performance and loss across different ranks.}
  \label{fig:combined}
\end{figure}

Interestingly, a rank-32 matrix—roughly an 11× compression in total parameters—achieves nearly 95\% of the teacher’s accuracy. This strongly suggests that for this task, during inference, we can scale down parameters significantly without sacrificing much performance. 

The early saturation effect shown by the effective knee at around $r^\star \approx 30.98$ also implies that most of the learning happens in these lower-rank regions. The large accuracy jumps between close ranks further support the hypothesis that most of the representational gain is concentrated in this region, after which the model mainly adds redundancy rather than new information.

\section{Ablation Studies}

We compared geometric distillation with pure distillation \citep{hinton2015distilling}, as well as MSE + cosine loss applied only to the output logits. Our results showed:

\begin{itemize}
  \item \textbf{Geometric Distillation}: 69.57\%
  \item \textbf{Logit MSE + Cosine}: 68.57\%
  \item \textbf{Pure KD} ($\alpha=0.9, T=4$): 61.24\%
  \item \textbf{Pure KD} ($\alpha=0.5, T=4$): 60.49\%
  \item \textbf{Pure KD} ($\alpha=0.9, T=2$): 56.91\%
\end{itemize}

Geometric distillation performed noticeably better than pure distillation across all three variants. However, using MSE + cosine loss only on the logits performed within about 1\% of full geometric distillation. This suggests that the method can be simplified—matching only the output logits while still achieving nearly identical results.

\begin{figure}[H]
  \centering
  \includegraphics[width=0.65\linewidth]{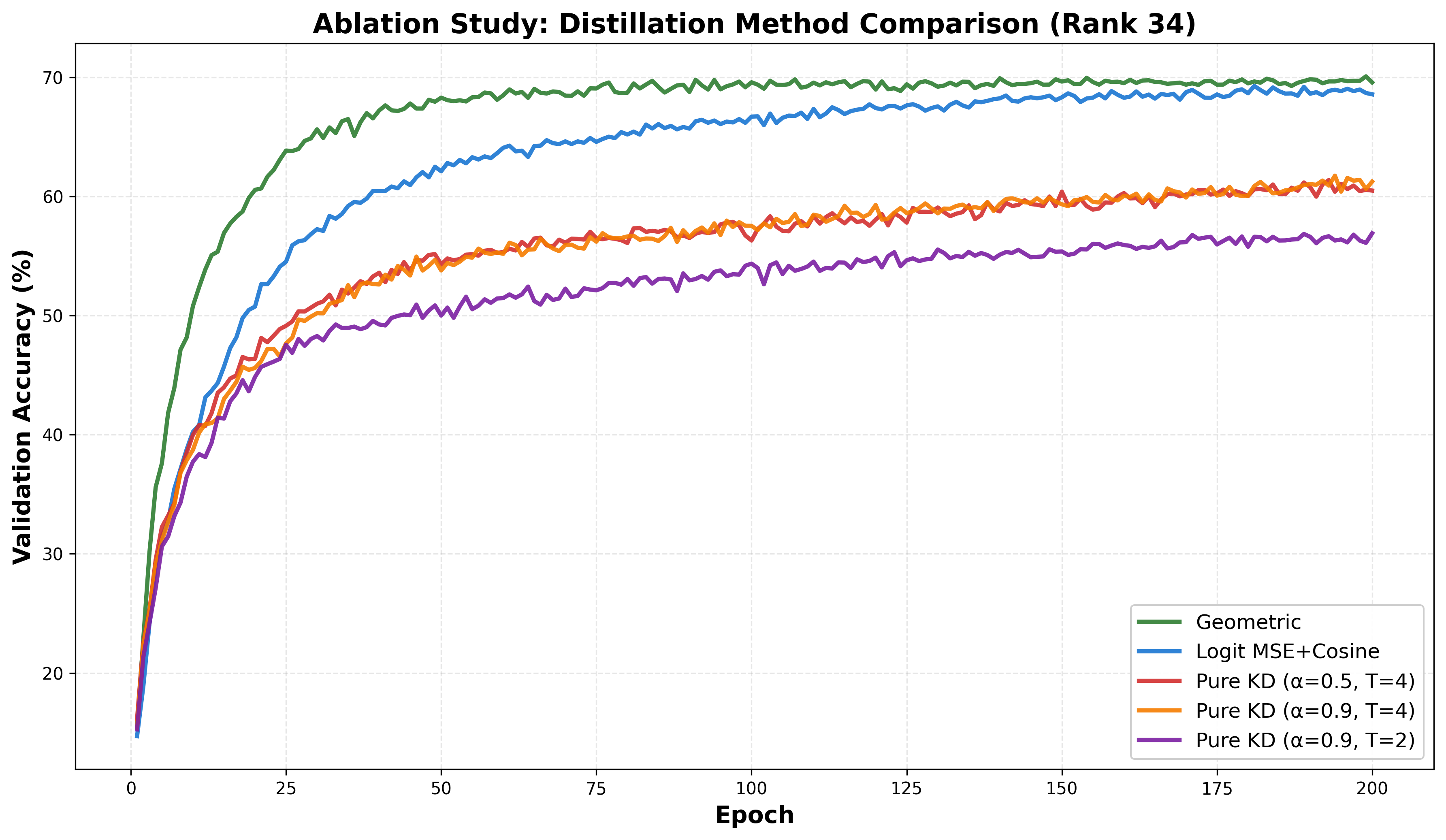}
  \caption{Factorized model performance across different distillation methods.}
  \label{fig:ablation}
\end{figure}

\section{Discussion}

Our results support the hypothesis that, on CIFAR-100, ViT-B/32 admits a low-rank representation at inference time. In particular, a rank-$32$ factorization yields $\sim$11$\times$ parameter compression (for the linear blocks) while attaining $\approx 94.7\%$ of the teacher’s accuracy. This suggests that substantial parameter down-scaling is feasible with modest loss, and that most of the utility concentrates in a relatively narrow rank band around the knee.

\paragraph{Where the gains concentrate.}
The observed early saturation (effective knee at $r^\star \approx 31$) indicates that marginal gains accumulate quickly in the low-rank regime and then taper off. The steep local slope of the accuracy–rank curve near $r^\star$ implies that representational capacity increases most rapidly in this band, consistent with a low-dimensional manifold view of the learned function.

\paragraph{Practical implications.}
For deployment scenarios with tight memory or latency budgets, sweeping a small grid of ranks around the knee (e.g., $r \in \{24,32,40\}$) can identify a favorable accuracy–efficiency trade-off without exhaustive search. Because the knee is stable under our protocol, this sweep is inexpensive and actionable.

\paragraph{Limitations.}
Our study is narrow by design(due to compute constraints): one architecture (ViT-B/32), one dataset (CIFAR-100), and a single seed due to compute constraints. We also focus on factorizing linear blocks; attention vs.\ MLP contributions are not disentangled in the main result. Finally, accuracy is the sole target metric; calibration and robustness are not evaluated.

\paragraph{A simple organizing model.}
Let the effective-rank region be governed by a task–model map
\[
\mathcal{E}_r \;=\; F(\text{architecture},\ \text{model size},\ \text{dataset},\ \text{dataset size})\,.
\]
In our case,
\[
\mathcal{E}_r \;=\; F(\text{ViT},\ \text{B/32},\ \text{CIFAR-100},\ 60{,}000)\,.
\]
Having a measurement protocol for $\mathcal{E}_r$ enables controlled studies that vary one factor at a time (e.g., B/16 vs.\ B/32; CIFAR-10 vs.\ CIFAR-100; or scaling the sample size) to chart how the region and knee shift.

\paragraph{Future work.}
Future directions include: (i) running multi-seed repeats and reporting confidence intervals for both the knee location and the width of the efficient region; (ii) cross-architecture evaluations (e.g., ConvNets, ViT variants, MLP-Mixers) and additional domains (e.g., NLP) to probe the universality of the effect; (iii) decomposing factorization targets (MLP-only vs.\ attention-only vs.\ joint) and reporting FLOPs and latency in addition to parameter counts; and (iv) going beyond accuracy to assess calibration and distribution shift, testing whether low-rank students inherit the teacher’s robustness.

\noindent\textbf{Code.} Implementation and experiments are available at
\url{https://github.com/LiyuZer/Geometric_Distillation}.

\section{Conclusion}
This work contributes toward understanding the intrinsic dimensionality of deep neural networks by providing a practical framework for estimating their effective rank. Our goal is to offer an algorithmic and automated way to identify the effective rank region---the range of ranks capturing a model’s intrinsic representational capacity.

By treating rank as a measurable property of learned representations, we enable systematic exploration of how dimensionality relates to model architecture, dataset size, and overall capacity. Such a framework opens the door to large-scale, automated analyses that can reveal deeper structure in what networks actually learn, offering predictive insight into the geometry and efficiency of their internal manifolds.

\bibliographystyle{plainnat}
\bibliography{refs}
\end{document}